\documentclass[letterpaper, 10 pt, conference]{ieeeconf}

\IEEEoverridecommandlockouts                              % This command is only needed if
\usepackage{refstyle}
\usepackage{hyperref}
\usepackage{graphics}
\usepackage{upgreek}
\usepackage{amssymb}
\usepackage{amsfonts}
\usepackage{amsmath}

\usepackage{xcolor}
\usepackage{subfigure}
\usepackage{todonotes}
\usepackage{mathtools}
\usepackage{cite}
\usepackage{balance}
\usepackage{mathrsfs}
\usepackage{lipsum}
\usepackage{multicol}
\usepackage{soul}
\usepackage{siunitx}

\usepackage{algorithm,algpseudocode}
\usepackage{booktabs}
\usepackage{optidef}
\usepackage{lineno}

\title{\LARGE \bf Physics-Encoded Graph Neural Networks\\ for Deformation Prediction under Contact
}

\author{Mahdi Saleh$^{1}$, Michael Sommersperger$^{1}$, Nassir Navab$^{2}$, Federico Tombari$^{3}$%
\thanks{Corresponding author: Mahdi Saleh (m.saleh@tum.de)}
\thanks{$^{1}$ M. Saleh, M. Sommersperger, are with Department of Computer Science, Technische Universit\"{a}t M\"{u}nchen, M\"{u}nchen 85748 Germany. 
}%
\thanks{$^{2}$N. Navab is a full professor and head of the Chair for Computer Aided Medical Procedures \&
Augmented Reality, Technical University of Munich, 85748 Munich, Germany.}%
\thanks{$^{3}$Federico Tombari is with the Faculty of Computer Science, Technische
Universität München, 85748 Garching bei München, Germany, and also with
the Google, 8002 Zurich, Switzerland (e-mail: tombari@in.tum.de).}%
}

\DeclarePairedDelimiterX{\norm}[1]{\lVert}{\rVert}{#1}

\definecolor{red}{rgb}{1,0,0}

\begin{document}

\maketitle

\renewcommand{\thefootnote}{\fnsymbol{footnote}} % Change footnote symbol for the abstract
\footnotetext[2]{Project page: \url{https://mahdi-slh.github.io/deform}} % Custom footnote with URL

%\thispagestyle{empty}

%\pagestyle{empty}

%%%%%%%%%%%%%%%%%%%%%%%%%%%%%%%%%%%%%%%%%%%%%%%%%%%%%%%%%%%%%%%%%%%%%%%%%%%%%%%%
\renewcommand{\thefootnote}{\fnsymbol{footnote}} % Change the footnote symbols to \fnsymbol for this part

\begin{abstract}
In robotics, it's crucial to understand object deformation during tactile interactions. A precise understanding of deformation can elevate robotic simulations and have broad implications across different industries. We introduce a method using Physics-Encoded Graph Neural Networks (GNNs) for such predictions. Similar to robotic grasping and manipulation scenarios, we focus on modeling the dynamics between a rigid mesh contacting a deformable mesh under external forces. Our approach represents both the soft body and the rigid body within graph structures, where nodes hold the physical states of the meshes. We also incorporate cross-attention mechanisms to capture the interplay between the objects. By jointly learning geometry and physics, our model reconstructs consistent and detailed deformations. We've made our code and dataset public to advance research in robotic simulation and grasping.
\footnotemark[2]
\end{abstract}

\renewcommand{\thefootnote}{\arabic{footnote}} % Reset back to arabic numbering after the abstract

\bstctlcite{IEEEexample:BSTcontrol}


\section{Introduction}
Grasping is a cornerstone of robotic manipulation, and accurate deformation prediction is pivotal for a stable and safe grip, especially with delicate or flexible objects. Miscalculations can result in improper handling, dropped items, or even damage. This precision becomes even more crucial in specialized fields like medical surgeries. Specifically, surgical tools must interact with delicate structures during procedures involving soft tissues, requiring high precision to prevent unintentional harm and guarantee patient safety~\cite{dehghani2023robotic}.

\begin{figure}
  \includegraphics[width=0.45\textwidth]{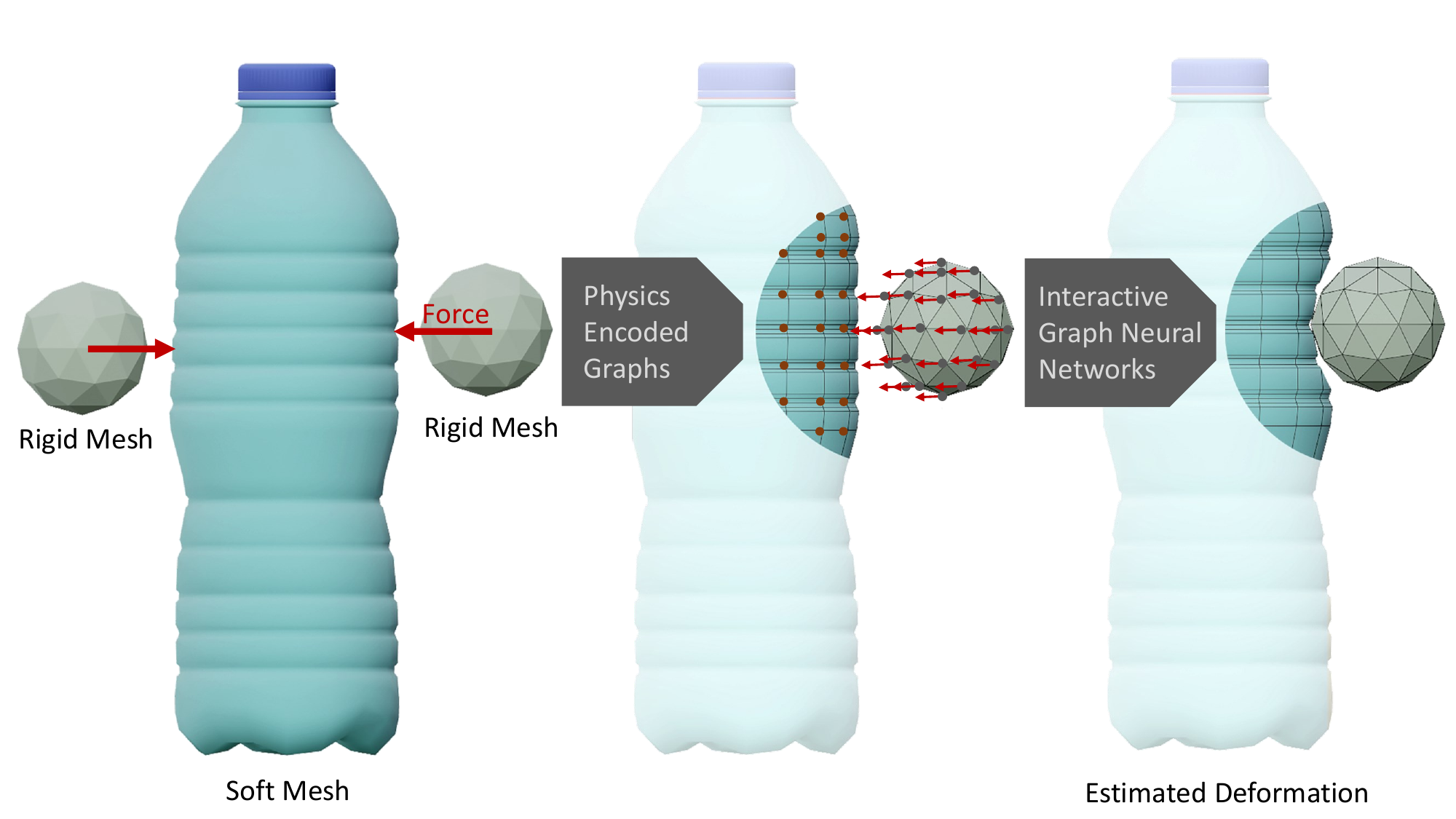}
  \caption{Given the input meshes of a soft and rigid body and their contact, we embed physical states into mesh graphs. Our network then processes the graphs to understand their interaction and predicts the deformation of the soft mesh. }
  \label{fig:teaser}
\end{figure}

Traditional techniques like Finite Element Methods (FEM)~\cite{clough1967analysis,applied1963analysis}, Boundary Element Methods (BEM)~\cite{hall1994boundary,xue2022scaled}, and Dual Mortar Methods~\cite{popp2010dual,popp2012dual} have been the widely used to predict deformation for a long time. However, they often struggle with complex, detailed meshes due to their computational intensity. Their optimization process can be tedious and time-consuming, especially for high-resolution models.

Moving beyond classical methods, particle-based methods\cite{gingold1977smoothed} offer a more flexible approach. Systems like Smoothed Particle Hydrodynamics (SPH)~\cite{liu2010smoothed} or Mass-Spring Systems \cite{provot1995deformation} approach deformation prediction by breaking down the object into a series of interconnected particles. These methods can handle more complex deformations but still can not ensure the accuracy of the predicted deformation on the original mesh.

Meanwhile, graph-based representations have gained traction in computer vision and graphics, especially in registration and deformable shape matching ~\cite{saleh2022bending,saleh2020graphite}. These methods can learn and match deformation patterns by viewing 3D shapes as a network of interconnected nodes. Graphs are particularly relevant for our problem formulation since we can encapsulate the physical characteristics of an object into the graph embeddings, ensuring a more detailed and structured representation.

Contrary to subsampled and holistic graph structures, we aim to use the original triangle mesh. By interpreting the connections of the triangle mesh as graph edges, our approach preserves the mesh structure, ensuring accurate deformation predictions without the additional steps typically associated with graph construction. We balance the reliability of classical methods and the flexibility of graph-based representations while reducing the computational costs. As depicted in Fig. \ref{fig:teaser}, we introduce a novel learning pipeline that embeds physics information, such as force features, into the graph nodes of both soft and rigid objects. Using GNNs, we capture the geometric shape features, and by using a cross-attention mechanism, we allow messages to pass between the soft and the rigid graphs. Finally, we reconstruct the post-contact mesh from the conditioned encoded features.

Our methodology can learn intricate surface deformations through a data-driven approach. The proposed network, adapted for our objectives, aims to bridge the gap between precision and computational efficiency. As the fields of robotic grasping and medical surgeries evolve, our integrated approach to learning physics and geometry together will significantly influence their direction.

We highlight our work's main contributions:

\begin{itemize}
    \item We curated a dataset of everyday objects with varied physical characteristics, associated metadata of contact, and deformations released together with the paper.
    \item We innovatively integrate the physical state of both soft and rigid bodies and their interaction into graph representations.
    \item Our novel GNN framework learns geometry and the underlying physics between objects, able to predict the distortion and deformations upon contact for different applications.
\end{itemize}

\section{Related work}\label{sec:Relatedwork}

\subsection{Classical Methods}
This section discusses classical and non-learning methods to estimate surface deformations. FEM provides a fundamental approach for estimating surface deformations by discretizing the problem domain into smaller elements. FEM has been widely used for deformation of plastic and other materials under stress or contact.~\cite{applied1963analysis,clough1967analysis,amor2021adaptive}. Boundary Element Methods (BEM), by contrast, focus on the boundary rather than the entire domain, optimizing computational efficiency for certain problems~\cite{hall1994boundary,xue2022scaled}. Mass-spring models are simplistic representations where a system is depicted using masses and springs to estimate deformations~\cite{provot1995deformation}.

Particle-based methods, such as Smoothed Particle Hydrodynamics (SPH), utilize particles to approximate deformations~\cite{gingold1977smoothed}. Dual mortar methods are mathematical formulations that allow for a more precise integration over contact interfaces~\cite{popp2010dual, popp2012dual}. However, they have complexity in implementation for many problems. Classical methods for surface deformation estimation provide a solid foundation and are used in various engineering applications. While classical methods offer robustness, learned techniques can promise efficiency and adaptability.

\subsection{Neural Physics}
Recent advances combine data-driven machine learning with physics, creating a paradigm known as Physics-Informed Machine Learning~\cite{karniadakis2021physics}. It enhances model performance by incorporating physical prior knowledge. While this interdisciplinary field has progressed, it remains under-explored in areas like robotic control~\cite{hao2022physics}. The work of \cite{chang2016compositional} showcases a Neural Physics Engine, a neural network architecture that breaks down object dynamics into pairwise interactions,  though it's mainly applied to simple two-dimensional dynamics. In a different approach, \cite{bekele2020physics} uses deep learning models trained on analytical solutions to predict deformations in poroelastic media.
Meanwhile, \cite{holden2019subspace} merges subspace simulation techniques with machine learning for efficient physics simulations, mainly focusing on dynamic cloths interacting with external objects. \cite{fulton2019latent} offers a framework for deformable solid dynamics using autoencoder neural networks, emphasizing a condensed representation through implicit integration. Additionally, \cite{okazaki2022physics} employs physics-informed neural networks to model crustal deformations. \cite{pfaff2020learning} and \cite{pfaff2020learning} leverage GNNs to predict cloth deformation and particle dynamics. Lastly, \cite{maurizi2022predicting} uses GNNs to understand material mechanics and predict different materials' structural deformations. 

\subsection{Graph Neural Networks}
Graph Neural Networks (GNNs) are extensively employed to process 3D geometry, including meshes and point clouds. Unlike point-based methods, which lack geometric intuitions \cite{qi2017pointnet}, GNNs leverage graphs to facilitate information exchange between neighboring points.
Graph Attention Networks (GATs) employ masked self-attentional layers tailored for graph data~\cite{velivckovic2017graph}. The multi-head attention concept can also combined with point and graph-based methods to attend to local and global features~\cite{zhao2021point,saleh2022cloudattention}. In parallel, Spectral CNN~\cite{yi2017syncspeccnn} facilitates 3D Shape Segmentation through weight sharing in a spectral domain through graph Laplacian eigenbases. EdgeConv dynamically refines edges across network layers, improving point cloud classification and segmentation tasks~\cite{wang2019dynamic}. Graphs have been then used for rigid~\cite{saleh2020graphite,saleh2022bending} and non rigid registration~\cite{saleh2022bending,yu2023rotation}. For instance, Graphite~\cite{saleh2020graphite} extracts keypoints and representations from a fixed spatial graph. Meanwhile, Bending Graphs uses mesh structures, offering a hierarchical approach to deformable shape matching~\cite{saleh2022bending}. Despite the capabilities of GNNs, much of the graph-centric and Geometric Deep Learning~\cite{bronstein2017geometric} research is restricted to correspondence estimation.

\subsection{Hand/Robot Object Interaction}
Robotic manipulation, once primarily focused on rigid objects, now frequently engages with deformable entities that alter shape upon contact~\cite{arriola2020modeling}. Such interactions make the underlying physics of deformation intricate and challenging to capture accurately. \cite{romero2021learning} proposes a subspace method for simulating dynamic deformations, merging subspace methods with learning-based techniques. Subsequent research by the same authors enhances these simulations with more detailed contact-driven deformations~\cite{romero2022contact}. In another study on deformable object simulations, \cite{leng2023dynamic} utilizes the hyperbolic space's unique properties for reconstructing hand and object meshes from images. Addressing deformations in a medical context, \cite{salehi2022physgnn} offers the PhysGNN framework, using GNNs to approximate soft tissue deformation during neurosurgical procedures.

While existing techniques offer significant insights into deformation prediction, the trade-offs between computational efficiency and precision become complex, especially in real-time scenarios. Recognizing this, our method aims to utilize the robustness of classical techniques, the adaptability of learning-based approaches, and the intricate modeling of GNNs, all integrated into a cohesive framework.

\section{Methodology}
\begin{figure*}[ht]
    \centering
  \includegraphics[width=0.85\linewidth]{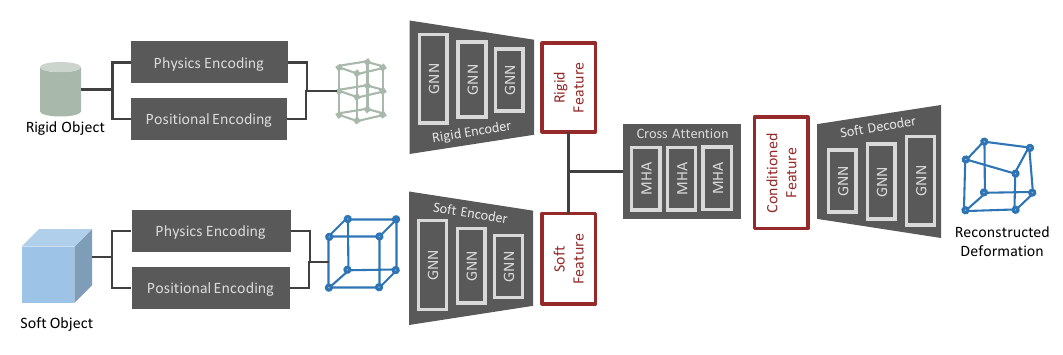}
  \caption{Our pipeline unfolds in three stages: First, we integrate the physics and positional encoding for both soft and rigid objects. Second, we engage in the encoder phase to extract features for each body. Finally, we employ multi-head attention to facilitate interactions between the features of soft and rigid objects, subsequently leveraging the conditioned feature to decode and reconstruct the deformed mesh or graph. }
  \label{fig:overview}
\end{figure*}

We present a compact pipeline to predict post-contact deformations by integrating physics encoding into GNNs. Our pipeline is also illustrated in Fig \ref{fig:overview}.

\subsection{Physics Encoding}
This subsection details the process of graph generation and representation. Given a triangle mesh, each vertex is interpreted as a node in the graph. While existing KNN graphs might work effectively for smaller datasets, querying nearest neighbors becomes computationally expensive as the number of vertices increases. Therefore, we produce three edges from each triangular mesh face, connecting each vertex. The information about neighboring vertices and how they relate to each other is preserved by referring to triangle meshes and can be easily manipulated or queried. 

Let \( P \in \mathbb{R}^3 \) denote the 3D positions of these vertices. These 3D coordinates, similar to the encoding in \cite{saleh2022bending}, are transformed into log-frequency representations to create more discriminative features. Moreover, we incorporate physics-related attributes for each node, such as force vectors, denoted by \( F \in \mathbb{R}^4 \). These initial attributes represent the normalized force vector and its magnitude.

The complete node feature for a given vertex then becomes a stack of its position information \( P \) and the associated force information \( F \). They are passed through multilayer perceptron (MLP) layers to encode these stacked features further. The resultant encoded node features serve as input for our graph neural network in the subsequent subsection.

\subsection{Network Architecture}

 Our network is designed to predict how points (or nodes) in a relaxed setting (the "soft" graph) change their positions when they come into contact with another entity, the so-called "rigid" graph. The architecture consists of layers of GNN for both the soft and rigid graphs, an interaction mechanism using Multi-Head Attention (MHA)~\cite{vaswani2017attention}, and finally, a decoder to reconstruct the object following the interaction.

 \textbf{GNN Encoders:} We take \(x_{\text{S}}\) and \(x_{\text{R}}\) as the initial state of our soft and rigid graphs, respectively. These features are processed and refined through GNN layers applied iteratively. Our GNN layers are based on TAGConv (Topology Adaptive Graph Convolution)~\cite{du2017topology}. TAGConv operates with varying topologies using adaptive, fixed-size filters for graph convolutions, ensuring more accurate geometric representations.

\textbf{Feature Interactions:} To establish interactions between the soft and rigid graphs, we leverage a cross-attention mechanism with multi-head attention~\cite{vaswani2017attention}. In this context, soft features from the resting graph and rigid features from the collider graph are input to the attention mechanism to capture intricate dependencies and relations between them. These interactions, illustrated in Fig. \ref{fig:overview}, allow us to condition the soft body features using the information from the rigid body, resulting in a richer, more representative feature set given the physics. 

\textbf{Soft Decoder: }Following this attention-based interaction, we obtain a conditioned feature representation that captures the mutual influence of the soft and rigid bodies on each other. These conditioned features represented as \( x_{\text{C}} \) are then processed by a graph-based decoder. This decoder refines and translates the conditioned features into the deformed mesh representation.

The output from the decoder, denoted as \( P_{\text{D}}\in \mathbb{R}^3 \), signifies the predicted deformed positions of nodes in the resting graph. The predicted graph output is directly used as the reconstructed mesh.

\subsection{Loss Functions}

\textbf{Mean Squared Error (MSE): }The MSE Loss measures the mean squared difference between the predicted positions ($P_{\text{D}}$) of nodes in the graph and the ground truth deformations of nodes ($P^*_{\text{D}}$). The MSE Loss is given by:

\begin{equation}
L_{mse} = \frac{1}{N_{\text{nodes}}} \sum_{i=1}^{N_{\text{nodes}}} \left| \text{P}_{\text{D}} - \text{P}^*_{\text{D}} \right|_2^2
\end{equation}

\textbf{Graph Consistency Regularization:} The Graph Consistency Regularization as a loss measures the consistency of gradient directions between the predicted graph ($D$) and the resting pose graph ($R$). Inspired by the graph Laplacian, which captures the intrinsic geometric structure of data, this term ensures that predicted deformations respect the underlying spatial relationships, mirroring physical constraints. The loss is computed as follows:

\begin{equation}
L_{G} = \frac{1}{N_{\text{edges}}} \sum_{i=1}^{N_{\text{edges}}} \left| \nabla(R) - \nabla(D) \right|_2
\end{equation}

Where:

$N_{\text{edges}}$ is the total number of edges in the graph.
$\nabla(R)$ represents the gradients of the resting graph ($R$) positions.
$\nabla(D)$ represents the gradients of the predicted graph ($D$) positions.
Graph Consistency Regularization is essential in preserving the object's structural integrity during deformation, ensuring that the model predictions adhere to realistic physical transformations.

Finally, the total loss, $L_{T}$, is computed as the weighted sum of the loss terms:

\begin{equation}
L_{T} =  L_{mse} + \lambda_{G} \cdot L_{G}
\end{equation}

The inclusion of $\lambda_{G}$ allows for the fine-tuning of the regularization effect, providing a mechanism to balance the influence of accurate position prediction with the preservation of structural consistency.

By tuning $\lambda_{G}$, we calibrate the model's focus between precision in spatial deformation prediction and adherence to realistic deformation behaviors, facilitating control over the model's generalization capabilities.
\section{Simulation and Datasets}
In this section, we detail our experimental setup and results. We start with an overview of our Simulation environment, designed for mesh integration and soft body simulation. Finally, we provide insights into our implementation details.

\subsection{Simulation System}
We select a subset of 8 objects from the Objaverse~\cite{objaverse} dataset and employ the Unity Engine~\cite{unity2023} for simulation. The soft body is positioned in the center of world coordinates, and Rigid objects are initialized with random translations and positions in space. Subsequently, we set random velocities and forces directed towards the soft body and observe the resulting collisions. For physics and deformation estimation, we utilize a unified particle physics~\cite{muller2007position} engine \footnote{\href{http://obi.virtualmethodstudio.com/}{http://obi.virtualmethodstudio.com/}}.

\begin{figure}[h]
    \centering
  \includegraphics[width=\linewidth]{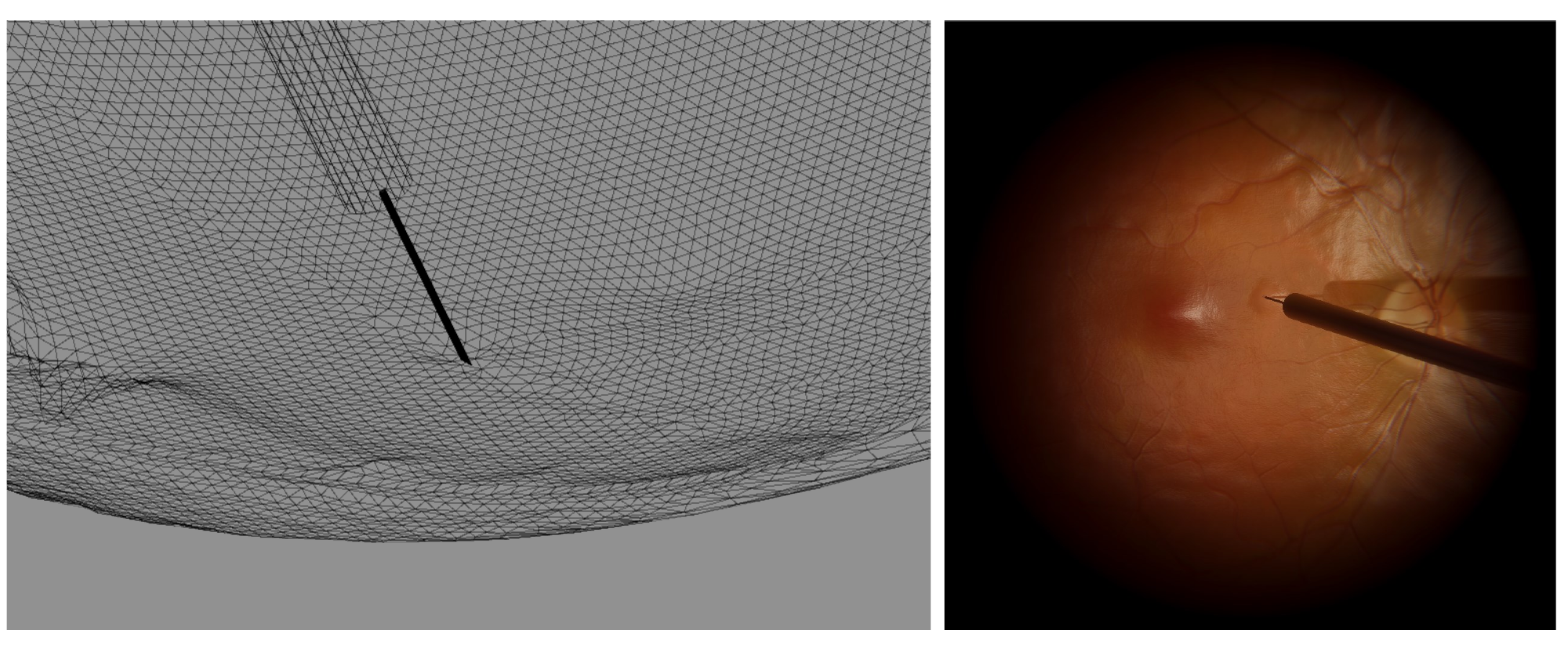}
  \caption{Images from the SynthesEyes simulation framework: On the left, the 3D mesh depicts the retina's deformation, while the right presents a 2D camera projection showcasing the intricate deformation patterns of the retina. We utilize and simulate these deformations for our training. }
  \label{fig:sim}
\end{figure}

\subsection{Everyday Deform Dataset}
\begin{figure}[h]
    \centering
  \includegraphics[width=\linewidth]{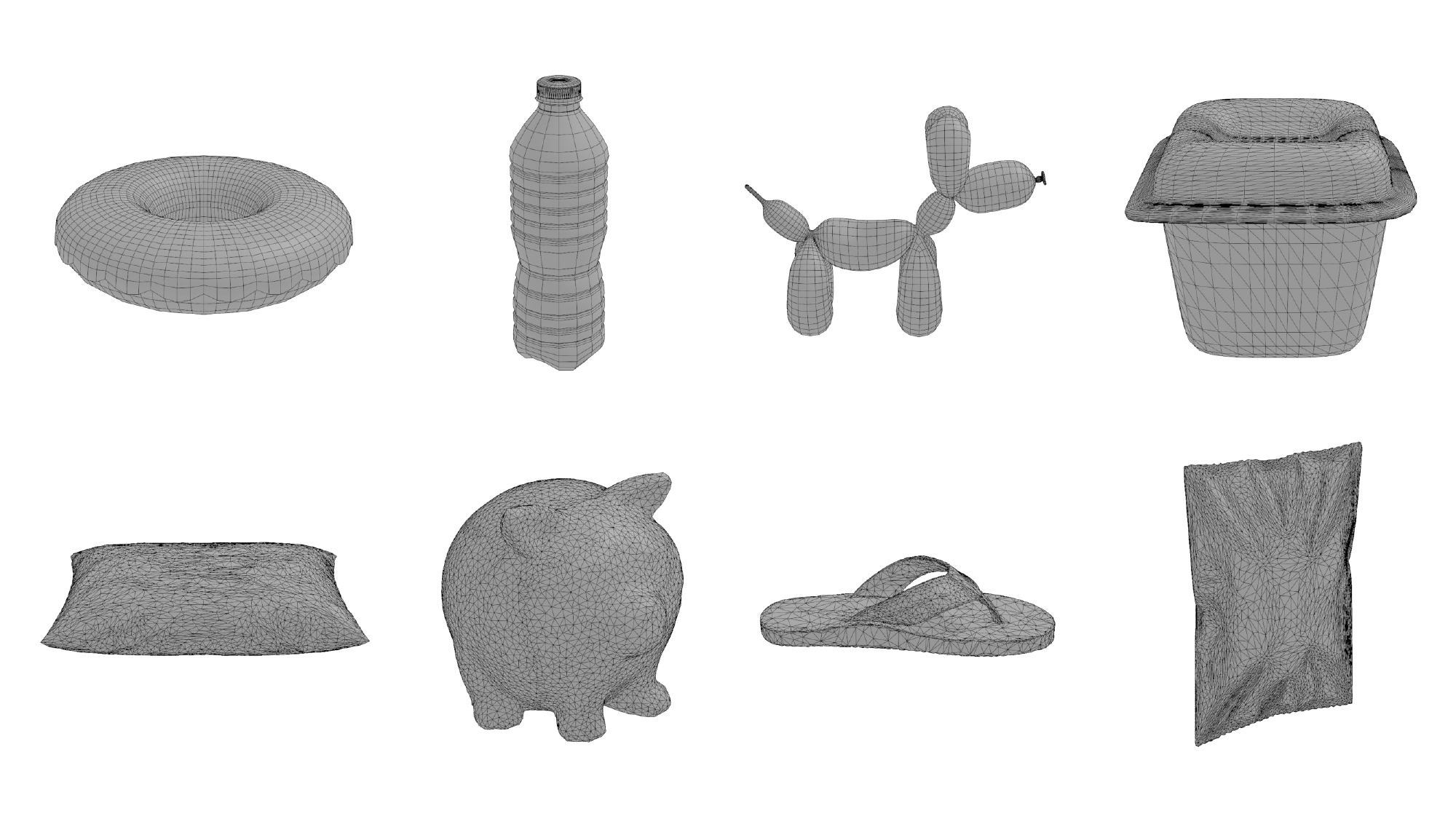}
  \caption{The 'Everyday Deform' dataset features eight commonly deformable objects, each with distinct mesh topologies, densities, and deformation characteristics. Using a simulator, each object undergoes random collisions with rigid bodies. The resultant mesh and associated physics data are recorded and made available to the community. }
  \label{fig:dataset}
\end{figure}

Leveraging our simulation system, we generate a dataset comprising eight distinct objects, each with unique deformation and stiffness attributes, as shown in Fig. \ref{fig:dataset}. These objects exhibit varied mesh densities. While some are sourced through 3D scanning, others are modelled using 3D CAD software, leading to diverse mesh topologies. The dataset includes the initial resting pose of each object and its deformation across 1.6k distinct contact scenarios. Each object has a fixed, manually determined stiffness and dampness. For every contact, we store the poses of the interacting entities, force vectors, velocities, and coordinates of contact points. In total, our dataset features around 11k meshes. We divided the dataset into an 80\%  training split and a 20\% test split.

\subsection{Retina Dataset}

Diving deeper into specialized applications, we create a dataset on the intricate task of reconstructing the retina surface mesh. Here, our primary objective is to model the deformation pattern of the retina's surface mesh upon its interaction with a surgical needle. Like the Everyday Deform dataset, we initiated our process by randomly positioning a needle around the retina surface. Here, we use custom software for vitreoretinal surgery simulation\footnote{\href{http://syntheseyes.de}{http://syntheseyes.de}}. We observed and recorded the subsequent deformations by applying varying degrees of force. This allowed us to simulate the complex physics involved, considering the known characteristics of the retina's delicate structure. This dataset consists of 1.5k meshes in different deformations upon contact. Fig. \ref{fig:sim} shows a view of the surgical tool interaction and rendering in the simulation software.

\subsection{Implementation Details}
We've constructed our mesh and deformation simulations within the Unity Engine~\cite{unity2023} framework. Post simulation, for the processing and learning phases, we integrated the PyTorch library for deep learning operations. For operations specific to graphs and mesh processing, we used Open3D and PyTorch Geometric libraries. 

Hardware-wise, our system is powered by a Nvidia Titan RTX GPU and an Intel(R) Core(TM) i7-7820X CPU.
For the training phase, we set our batch size at 4. The meshes we processed ranged from 2-6k vertices, dynamic with the object in the pipeline. Notably, we process the full meshes, avoiding sampling techniques. For the retina dataset, our graphs incorporated a detailed 64-node structure to reflect the granularity of the deformation we aimed to capture. Our input features to GNNs are of size 21 and 25 for soft and rigid graphs, and the intermediate encoded feature size is 256.

\section{Results and Discussions}

To assess the accuracy and efficacy of our proposed method, we adopt specific evaluation metrics and conduct a series of experiments. Our primary evaluation metric is MSE error. MSE calculates the average squared Euclidean distance between the nodes of the predicted deformation and the corresponding nodes in the ground truth. Moreover, we calculate the Mean Absolute Error (MAE) as the average absolute difference between the predicted deformation and the ground truth. Finally, Graph Consistency evaluates the consistency and smoothness of the predicted deformation relative to the ground truth.

\subsection{Quantitative Evaluation} 
\textbf{Experiment 1: Object Contact.} 
In our initial experiment, we assess the efficacy of our methodology on the Everyday Deform dataset. Our evaluation strategy is twofold. First, as detailed in Table \ref{table:everyday_objects}, we train a single network per object for 40 epochs. These objects, characterized by diverse densities and physical characteristics, are effectively learned by our model to estimate their full-body deformation. The Dog demonstrates optimal shape consistency while achieving the lowest MSE error. On the other hand, despite the Bottle's high deformability, it performs the best in terms of MAE error.

\begin{table}[h!]
\centering
\begin{tabular}{c|c|c|c|c}
\hline
\textbf{Object} & \textbf{Size} & \textbf{Consistency} & \textbf{MAE} & \textbf{MSE} \\
\hline
Bottle & 3927 & 0.0015 & 3.02e-05 & 3.74e-05\\
Cat & 5002 & 0.0023 & 0.0001 & 0.0002 \\
Dog & 4612 & 0.0010 & 2.92e-05 & 1.87e-05 \\
Donut & 2560 & 0.0012 & \textbf{7.38e-06} & 1.89e-05 \\
Doritos & 5002 & 0.0013 & 4.27e-05 & 2.94e-05 \\
Flipflop & 4993 & \textbf{0.0009} & 2.63e-05 & \textbf{1.35e-05} \\
Pillow & 4002 & 0.0019 & 3.87e-05 & 6.02e-05 \\
\hline
\textbf{Average} & - & 0.0013 & 3.79e-05 & 2.57e-05 \\
\hline
\end{tabular}
\caption{Comparison of deformation prediction for everyday objects. Each network is trained separately on the full mesh.}
\label{table:everyday_objects}
\end{table}

In another experiment, we evaluated our model's performance, trained using the complete training set on the closest mesh patch comprising 1024 vertices. The results of this evaluation are outlined in Table \ref{table:together}. Even though this dataset encompasses a spectrum of physical attributes and topologies, our rich representation facilitates consistent performance on unseen deformation instances.

\begin{table}[h!]
\centering
\begin{tabular}{c|c|c|c}
\hline
\textbf{Object} & \textbf{Consistency} & \textbf{MAE} & \textbf{MSE} \\
\hline
Bottle & 0.0045 & \textbf{6.11e-05} & 0.000143 \\
Box & 0.0034 & 5.75e-05 & 0.000105 \\
Cat & 0.0050 & 0.000111 & 0.000312 \\
Dog & \textbf{0.0013} & 7.02e-05 & \textbf{2.99e-05} \\
Donut & 0.0030 & 1.89e-05 & 7.44e-05 \\
Doritos & 0.0027 & 8.71e-05 & 7.75e-05 \\
Flipflop & 0.0019 & 6.31e-05 & 6.44e-05 \\
Pillow & 0.0037 & 7.74e-05 & 0.000158 \\
\hline
\textbf{Average} & 0.0032 & 6.69e-05 & 0.000119 \\
\hline
\end{tabular}
\caption{Comparison of deformation prediction error for random local objects. A single network is trained for all patches.}
\label{table:together}
\end{table}

\textbf{Experiment 2: Retina Tool Interaction.} 
Our second experiment focuses on the challenging task of retina surface mesh reconstruction. Specifically, we aim to predict the deformation of the retina surface mesh when it comes into contact with a surgical needle.

\begin{table}[h!]
\centering
\begin{tabular}{c|c|c|c}
\hline
\ & \textbf{MAE (µm)} & \textbf{MSE (µm)} & \textbf{Variance (µm)} \\
\hline
Retina & 57.326 & 10.88 & 0.00010348 \\
\hline
\end{tabular}
\caption{Error metrics for retina tool deformation estimation, in micrometers.}
\label{table:retina_interactions}
\end{table}

Table \ref{table:retina_interactions} summarizes this experiment's results, demonstrating our methodology's capability to accurately reconstruct the retina's local surface under varying force conditions from surgical tools.

\subsection{Qualitative Evaluation} 
\begin{figure}[h]
    \centering
  \includegraphics[width=\linewidth]{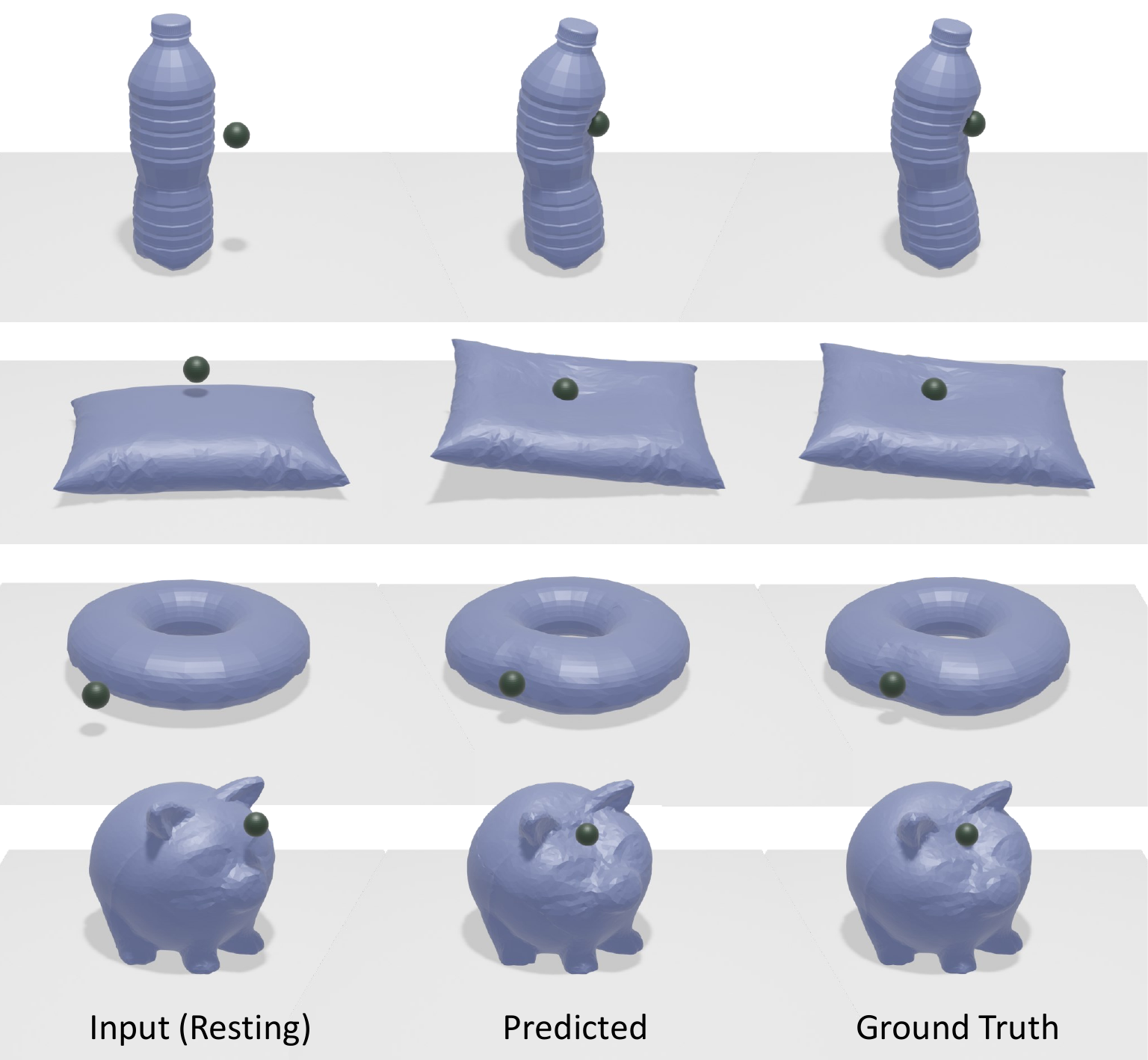}
  \caption{The figure presents our quantitative results calculated from the Everyday Deform dataset. The resting pose is visualized on the right, with the rigid object in its initial position. At the center, we illustrate the deformation prediction from our network. The left images depict the ground truth deformation present in the dataset. Our method captures fine surface deformation rigid motion and even reflects interaction with the ground floor.}
  \label{fig:result_everyday}
\end{figure}

\begin{figure}[h]
    \centering
  \includegraphics[width=\linewidth]{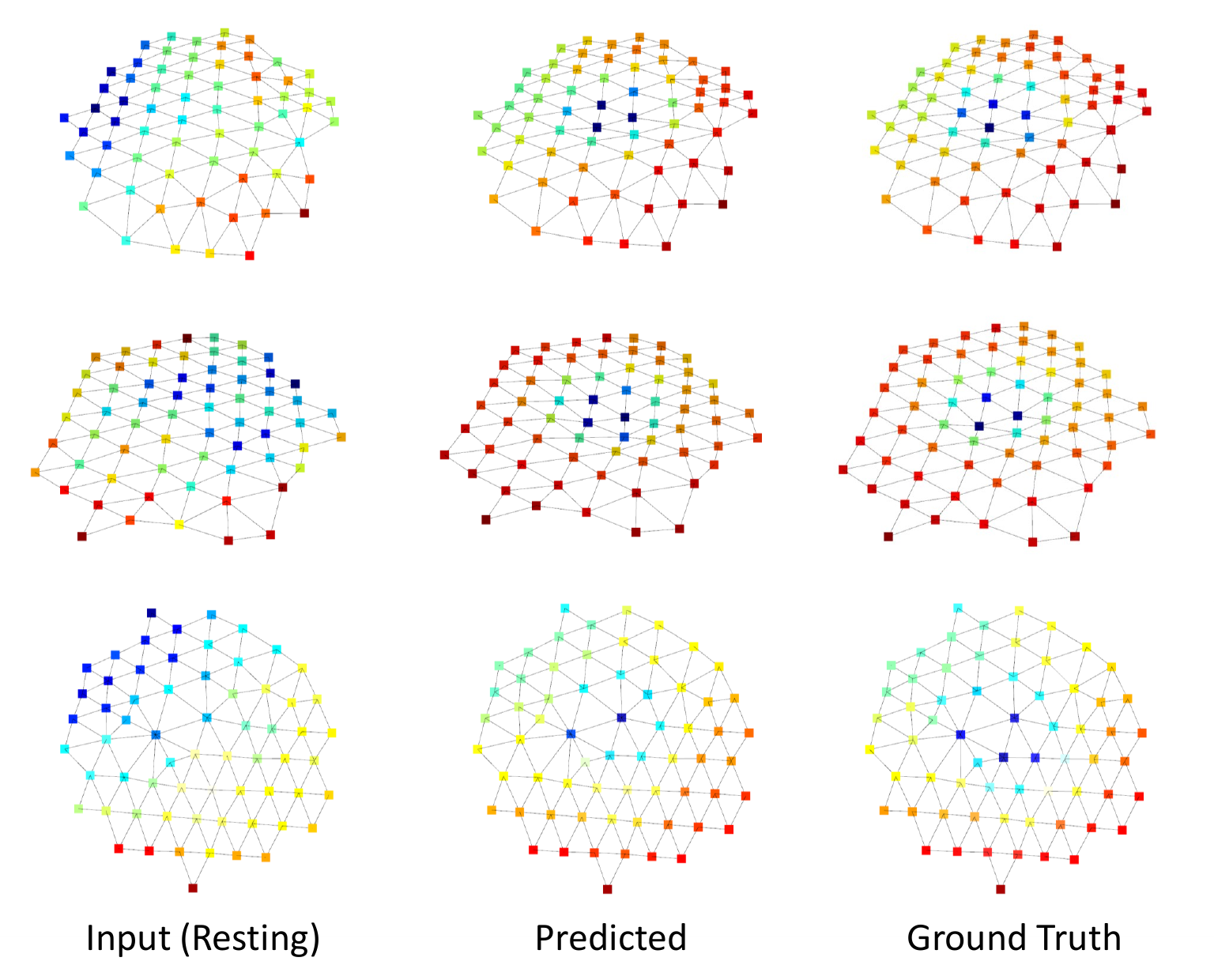}
  \caption{The figure provides a visualization of the retina mesh graph: the input is displayed on the left, the predicted deformations in the middle, and the ground truth on the right. Predicted vertex normal vectors are depicted, using color mappings to highlight the curvature. }
  \label{fig:result_retina}
\end{figure}

Visualization results in Figures \ref{fig:result_everyday} and \ref{fig:result_retina} present an intuitive perspective into the model's predictions. Our model can clearly predict fine details, be it in terms of surface deformations or interactions with surrounding environments. The comparative visualizations, especially between predicted and ground truth deformations, highlight the minimal discrepancy, reaffirming the quantitative findings.

\subsection{Ablation Studies}

To better understand the intricacies of our model and the influence of distinct encodings, we executed ablation studies, the results of which are presented in Table \ref{table:ablation}. By selectively activating the positional and physics encodings — individually and in combination — we could assess their individual and combined effects on prediction accuracy. The results after ten epochs indicate that the highest accuracy in our node's initial state is achieved when both geometric and physics information are simultaneously encoded.

\begin{table}[h!]
\centering
\begin{tabular}{c|c|c|c|c}
\hline
& \textbf{Positional} & \textbf{Physics} & \textbf{MAE} & \textbf{MSE}  \\
\hline
\textbf{Only Physics} & & \checkmark & 1.05e-05 & 3.119e-04 \\
\textbf{Only Positional} & \checkmark & & 1.14e-05 & 3.389e-04 \\
\textbf{Both} & \checkmark & \checkmark & \textbf{0.90e-05} & \textbf{2.695e-04} \\
\hline
\end{tabular}
\caption{Ablation study on different encodings using MAE, MSE, with active encodings indicated by a checkmark. The study shows both encodings contribute to the final predictions.}
\label{table:ablation}
\end{table}

\subsection{Runtime}

Practical applications require not just accuracy but also efficiency. Our runtime evaluations, presented in Table \ref{table:runtime}, suggest that our method is computationally efficient, rendering it suitable for real-time applications. The table shows latency for graph and data preparation and network inference pass. This is particularly crucial for surgical systems and simulations where timely and quick predictions are necessary. 

\begin{table}[h!]
\centering
\begin{tabular}{c|c|c|c}
\hline
\textbf{Dataset} & \textbf{Size} & \textbf{Data (ms)} & \textbf{Network (ms)}\\
\hline
Retina & 64 & 0.992 & 2.978 \\
Everyday Deform & 1024 & 1.661 & 4.261 \\
\hline
\end{tabular}
\caption{Runtime evaluation per batch (of 4 samples) on different datasets, split into data loading and graph generation, and network pass and estimation.}
\label{table:runtime}
\end{table}

\subsection{Future Works}
We plan to tackle a few key areas to improve our method in our upcoming work. One idea is to use time series analysis. Our method currently looks at deformation at a single point in time, but with time series, we could track how objects change over a more extended period. This could be useful in tasks like robotic planning. Another thing we're excited about is using Neural Ordinary Differential Equations (Neural ODEs). These tools are great at showing continuous changes, which could give a smoother picture of how things like retina tools or soft objects interact.

Additionally, our current dataset doesn't account for different deformable parts of an object. One can pay more attention to this in the future so the simulations are closer to true-to-life objects, especially the ones with mixed materials. And finally, instead of the focus on soft-to-rigid interaction, one can build on top of our pipeline to learn how soft objects interact with each other. These updates will make our contribution even better and more in tune with real-world situations.
\section{Conclusion}
\label{sec:conclusion}
This work bridges the gap between computational geometry and Physics-Informed Deep Learning, utilizing a data-driven approach to model object interactions. We process mesh graphs for soft and rigid shapes, integrating physical and geometric data. By leveraging cross-attention mechanisms, we capture fine-grained physical dynamics. Subsequently, we reconstruct post-contact deformations under various conditions. Supplemented by a new dataset featuring diverse everyday objects, our approach presents a practical and efficient solution for simulated robotics, grasping, surgical tool interactions, and other contact modeling and robotic manipulation applications.

\section{Acknowledgment}
The authors wish to thank SynthesEyes for providing the excellent simulation setup for the data generation.
\clearpage

\bibliographystyle{IEEEtran}

\newpage
{
\bibliography{ref}
}
\end{document}